# ON GRANULAR KNOWLEDGE STRUCTURES

Yi Zeng[1], Ning Zhong[1, 2]

[1] International WIC Institute, Beijing University of Technology
Beijing, 100124, P.R. China.  Email: yzeng@emails.bjut.edu.cn
URL: http://www.iwici.org/~yizeng
[2] Department of Life Science and Informatics, Maebashi Institute of Technology
Maebashi-City, 371-0816, Japan.  Email: zhong@maebashi-it.ac.jp
URL: http://kis-lab.com/zhong

## Abstract

Knowledge plays a central role in human and artificial intelligence. One of the key characteristics of knowledge is its structured organization. Knowledge can be and should be presented in multiple levels and multiple views to meet people's needs in different levels of granularities and from different perspectives. In this paper, we stand on the view point of granular computing and provide our understanding on multi-level and multi-view of knowledge through granular knowledge structures (GKS). Representation of granular knowledge structures, operations for building granular knowledge structures and how to use them are investigated. As an illustration, we provide some examples through results from an analysis of proceeding papers. Results show that granular knowledge structures could help users get better understanding of the knowledge source from set theoretical, logical and visual point of views. One may consider using them to meet specific needs or solve certain kinds of problems.

## 1 Introduction

Knowledge plays an important role in both human and artificial intelligence. Minsky argues that small fragment of knowledge without connecting with a large knowledge structure of human thoughts is meaningless [8], which emphasize the importance of knowledge structure in knowledge representation. In addition, some studies show that visualized structure of knowledge can help people get better understanding when they are facing great volume of information [5, 18]. Many related studies claim that knowledge and its structure can be and should be presented in multiple levels and multiple views [1, 8, 18, 19]. Multi-level representation presents knowledge in a hierarchical way, which satisfies people's needs in different level of granularities. While multi-view representation provides different understandings of the same knowledge source from different viewpoints, which may help people understand the knowledge source from different perspectives [8, 19]. The triarchic model of Granular Computing emphasize on structured thinking, structured problem solving and structured information processing [19], in which structure plays a central role. In other words, both knowledge and granular computing emphasize on the importance of structures. Hence, we stand on the view point of granular computing and provide our understanding on multi-level and multi-view of knowledge through granular knowledge structures. Literature is a source of recorded knowledge, and the amount of it is growing rapidly. One can get the implicit knowledge and its structure through reading, analyzing and learning those literatures, which can also be done by machine through some analysis method [10, 11, 18]. Granular Knowledge Structure can be used as an intelligent, knowledgeable way for organizing literatures, which can help people understand those literatures. As a demonstration of our proposed idea, we analyze some proceeding papers from some Rough Sets related conferences. The examples show that granular knowledge structures not only can give interpretations of the knowledge source from multiple levels and multiple views, but also can support more intuitive understanding.

## 2 Defining Granular Knowledge Structures

Granule is a high level abstraction of an element, or a set of elements with concrete meanings. A basic granule is considered to be indivisible or there is no need to divide [19]. Elements in a granule are drawn together by indistinguishability, similarity, proximity, functionality, etc [6]. Different fields have different units corresponding to granules [19]. Knowledge and its structures can be better formulated and learned if they are organized based on concepts and relations among them [3, 4, 9]. Hence, in the context of granular computing, we provide a way of representing granular knowledge structures based on concept granules. Concept granule can be defined based on an information table [16], and it serves as a basis for defining granular knowledge structures.

**Definition 1**: (Information Table) Formally, an information table can be represented as follows [16, 20]:
$$T = (U, At, V_a, R_a, I_a), \qquad (1)$$
where $a \in At$, $U$ represents a finite nonempty set of Yi Zeng, Ning Zhong. On Granular knowledge Structures, In: Progress of Advanced Intelligence: Proceedings of 2008 International Conference on Advanced Intelligence, Posts and Telecommunications Press, Beijing, China, October 18-22, 2008, 28-33.

objects, $At$ represents a finite nonempty set of attributes, $Va$ represents a nonempty set of values for $a \in At$, $Ra$ represents a set of binary relations on $Va$, $Ia: U \rightarrow Va$ is an information function for $a \in At$ [16, 20].

The logic language $\mathcal{L}$ for granular computing [20], which is an extension of a decision logic language used by Pawlak [14], is the basis for granule representation and reasoning with granules.

**Definition 2**: (Atomic Formula) In the context of information table, objects can be grouped together as granules based on formulas. A formula (denoted as $\phi$) can be an atomic formula or a combination of atomic formulas (Through logical operations defined in the language $\mathcal{L}$) [20]. An atomic formula can be represented as [16]:

$$(a, r, v) \quad (2)$$

where $r \in Ra$ denotes a binary relation between an attribute ($a \in At$) and the corresponding attribute value ($v \in Va$) [16]. Possible binary relations are equality relation, equivalence relation, similarity relation, etc [15, 17].

**Definition 3**: (Concept Granule) A concept is considered to be the basic unit of human thoughts and the component of knowledge [4], and it can be conveniently represented by its intension and extension [7]. In the context of granular computing, a concept can be interpreted as a concept granule, and it can be formally defined as [15]:

$$(\phi, m(\phi)) \quad (3)$$

where the formula $\phi$ represents the intension of a concept granule, while $m(\phi)$ is the set of objects satisfying $\phi$ and represents the extension of a concept granule [15].

**Example 1**: We draw an example of information table from [18], as shown in Table 1, which is a partial analysis of papers in proceedings of RSFDGrC 2005 and RSKT 2006. Values in the column "Theory" represent Rough Sets related theories which appear in these papers, while values in the column "Application Domain" represent the related application domains that these papers refer to. Following is an example of a concept granule based on Table 1:

$$((Theory, =, FCA), m(Theory, =, FCA)) . \quad (4)$$

The intension of this concept granule is an atomic formula which represents that the theory contained in the papers is formal concept analysis. The extension of it is the set of papers which uses formal concept analysis.

In conceptualized world, concept granules are not isolated, and they are interconnected based on relations [4, 8]. In this paper, we introduce a basic binary relation that may help to build granular knowledge structures.

**Definition 4**: (Partial Ordered Relation) Since the extension of a concept granule corresponds to a set of elements satisfying its intension, a partial ordered relation on two concept granules can be defined based on set inclusion [13]:

$$(\phi, m(\phi)) \preceq (\varphi, m(\varphi)) \Leftrightarrow m(\phi) \subseteq m(\varphi) . \quad (5)$$

Partial ordered relation can be used to describe relations among sub-concept granules and super-concept granules. A concept granule $(\phi, m(\phi))$ is regarded to be a sub-concept granule of another concept granule $(\varphi, m(\varphi))$, or $(\varphi, m(\varphi))$ a super-concept granule of $(\phi, m(\phi))$ if $m(\phi) \subseteq m(\varphi)$ [16].

Table 1: A partial information table describing papers from proceedings of RSFDGrC 2005 and RSKT 2006 (from[18]).

| Paper | Initial Page | Theory | Application Domain | Discipline |
|---|---|---|---|---|
| No.05 | p1-94 | R-A | – | Rough Sets |
| No.12 | p1-345 | RFH | – | Rough Sets |
| No.25 | p2-342 | LR | MS | Rough Sets |
| No.21 | p2-263 | DR | IP | Rough Sets |
| No.29 | p2-383 | LR | BI | Rough Sets |
| No.97 | p3-522 | FCA | – | Rough Sets |
| No.30 | p2-430 | DR | BI | Rough Sets |

R-A: Rough-Algebra, LR: Logics and Reasoning, RFH: Rough-Fuzzy Hybridization, FCA: Formal Concept Analysis, DR: Data Reduction, MS: Medical Science, BI: Bioinformatics, IP: Image Processing, DT: Decision Table, RPA: Rough Probabilistic Approach, GC: Granular Computing, RA: Rough Approximation, IR: Information Retrieval, MS: Medical Science, IS: Information Security.

**Example 2**: Following is an example of partial ordered relation between two concept granules in Table 1:

$$((Theory, =, FCA), m(Theory, =, FCA)) \preceq$$
$$((Discipline, =, Rough\ Sets), m(Discipline, =, Rough\ Sets)).$$

Relations show how concept granules are connected to each other [4]. One may define other binary relations between concept granules. In the context of Artificial Intelligence and Cognitive Psychology, a composition of concepts and relations can be used to form a conceptual graph, which can be used to represent knowledge [1, 4, 8, 9]. From the view point of granular computing, we can use concept granules and relations among them to describe granular knowledge structures.

**Definition 5**: (Granular Knowledge Structure) A granular knowledge structure can be defined as follows:

$$GKS = (\{(\phi_n, m(\phi_n)) \mid n \in I^+\}, \{\mathcal{R}_i \mid i \in I^+\}), \quad (6)$$

where $\{\mathcal{R}_i \mid i \in I^+\}$ denotes the set of binary relations among the set of concept granules $\{(\phi_n, m(\phi_n)) \mid n \in I^+\}$, and $I^+$ denotes the set of positive integers. Each two concept granules on one binary relation form an ordered pair. Let $i, j \in I^+$, the ordered concept granule pair can be represented as $< (\phi_i, m(\phi_i)), (\phi_j, m(\phi_j)) >$, where $(\phi_i, m(\phi_i))$ and $(\phi_j, m(\phi_j))$ are two concept granules.

A granular knowledge structure emphasizes on how the concept granules are organized. If concept granules involved in the granular knowledge structure can be organized into levels, then the granular knowledge structure is a hierarchy composed of concept granules. Concept granules in the same level may share some commonalities. If they cannot be organized into levels, they may form a concept granule network. One can get intuitive understanding of knowledge through different granular knowledge structures from different views, which can be induced based on various operations.

## 3 Building Granular Knowledge Structures

In this section, we introduce some basic operations for producing granular knowledge structures in different views. Scientific literature is a source of recorded existing knowledge, which can be organized into different granular knowledge structures. Hence, we choose papers from two proceedings and build granular knowledge structures based on basic operations introduced in this section. Firstly, we introduce how to generate granular knowledge structures based on an information table, then we examine how to generate granular knowledge structures based on existing granular knowledge structures. An explanation of the meaning on each granular knowledge structure is provided. Through analyzing all papers in proceedings of RSFDGrC 2005 and RSKT 2006 through an information table, from which Table 1 is extracted, the set of values on attribute "Theory" and attribute "Application Domain" are provided:

$V_{Theory} = \{DT, RPA, R\text{-}A, LR, RFH, GC, RA, FCA, DR\}$,

$V_{ApplicationDomain} = \{IR, MS, IS, BI, IP\}$.

One can separately define two sets of concept granules based on equality relation on $At = \{Theory\}$ and $At = \{Application\ Domain\}$. These concept granules correspond to a group of papers which use these theories or are related to these application domains. Hence we can build some simple granular knowledge structures based on attributes and their values.

**Definition 6**: (Attribute-Value Structure) In an information table, let an attribute $a \in At$ and it has a corresponding set of attribute values, denoted as $\{V_a | a \in At\}$. One can generate a set of concept granules based on equality relations on attribute and attribute values. A more general concept granule, denoted as $(\rho, m(\rho))$, can be formed to include all concept granules induced by $(a, =, v)$, where $\rho$ is a disjunction of all $(a, =, v)$, for each $a \in At$. We can use the attribute name to label $(\rho, m(\rho))$ to show its meaning. Hence, for each $a \in At$, a concept granule in the form of $((a, =, v), m(a, =, v))$ has a partial ordered relation with $(\rho, m(\rho))$. And one can form an attribute-value structure through composing all the concept granules and partial ordered relations.

**Example 3**: As for $At = \{Theory\}$, papers which use at least one of the theories related to Rough Sets are grouped together to form the coarsest concept granule. Concept granules induced by equality relation on $At = \{Theory\}$ are sub-concept granules of the coarsest concept granules. These concept granules can form a granular knowledge structure based on partial ordered relations. One can produce a similar granular knowledge structure considering $At = \{Application\ Domain\}$. These two granular knowledge structures are shown in Figure 1. $[X]$ denotes a concept granule, and if the intention of it is an atomic formula, $X$ is the attribute value, and if the intention is the conjunction of all formula for each $a \in At$, $X$ is the attribute name. The line shows the partial ordered relations between concepts.

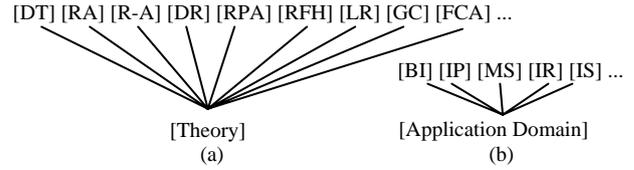

Fig. 1: Examples of attribute-value structure.

The concrete meaning of Figure 1(a) is that the current two proceedings cover nine theories related to Rough Sets. Figure 1(b) shows that the current two proceedings cover five application domains.

**Definition 7**: (Generalization Operation) Considering two concept granules $(\phi, m(\phi))$ and $(\varphi, m(\varphi))$, they may share an attribute-value pair which are the same. One may consider providing a more general concept granule as their super-concept granule (the attribute value can be used to label this concept granule), which can be used to describe this special relation. A new granular knowledge structure is produced.

**Example 4**: With respect to Figure 1(a) and Figure 1(b), the two concept granules [Theory] and [Application Domain] share the same attribute and attribute value $(Discipline, =, Rough\ Sets)$. We consider providing a more general concept granule [Rough Sets] as their super-concept granule. The new granular knowledge structure is shown in Figure 2, which shows an understanding of Rough Sets from two views, namely, related theories and application domains.

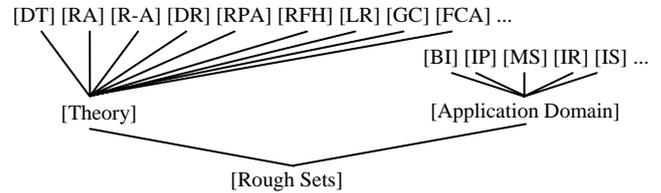

Fig. 2: Generalization operation on GKS.

**Definition 8**: (Union Operation) Let $(\phi, m(\phi))$ be a concept granule that two arbitrary granular knowledge structure (denoted as $GKS_1$, $GKS_2$) both have. In each granular knowledge structure, the concept $(\phi, m(\phi))$ can be composed of sub-concept granules which are one level finer than it. Relations among $(\phi, m(\phi))$ and its sub-concept granules can be denoted by relations among their extensions:

In $GKS_1: m(\phi) = m(\phi_1) \cup m(\phi_2) \cup ... \cup m(\phi_n)$,

In $GKS_2: m(\phi) = m(\phi_{n+1}) \cup m(\phi_{n+2}) \cup ... \cup m(\phi_{n+p})$,

where $n, p \in I^+$. A union operation can be used to make a union of these two sets of sub-concept granules, and induces a new granular knowledge structure. Relations among $(\phi, m(\phi))$ and its new set of sub-concept granules can be denoted by their corresponding extensions:

$$m(\phi) = m(\phi_1) \cup m(\phi_2) \cup ... \cup m(\phi_k), \qquad (7)$$

where $k \in I^+$. Notice that sub-concept granules which share the same intention need to be merged together to the same one. Their corresponding extensions are also grouped together as the extension of the new one. This operation helps to understand how a knowledge structure can be constantly evolving by merging related knowledge source.

**Example 5**: Figure 3(a) and Figure 3(b) are two granular knowledge structures considering related theories in proceedings of RSFDGrC 2005 and RSKT 2006. Since the bottom concept granules of these two structures are all [Theory], we can use union operation to obtain a unified structure, which provides a more complete description for the sub theories of Rough Sets, as shown in Figure 3(c).

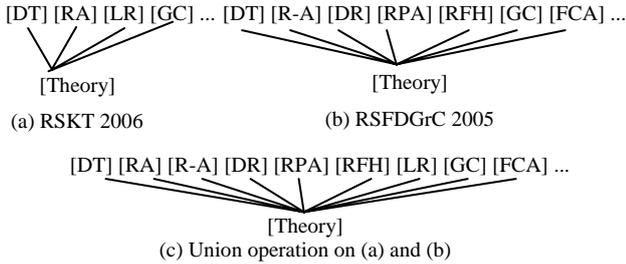

Fig. 3: Union operation on GKS.

**Definition 9**: (Intersection Operation) Considering two concept granules $(\phi, m(\phi))$ and $(\varphi, m(\varphi))$ from two granular knowledge structures, if and only if $\phi = \varphi$, an intersection operation is used to find the same sub granular knowledge structure induced by these two concept granules and their sub-concept granules. A sub granular knowledge structure is a partial structure from the whole granular knowledge structure. Since $\phi = \varphi$, the two concept granules are merged together. Except for this concept granule, extensions of other concept granules contained in the sub granular knowledge structure can be represented as:

$$\{m(\tau_k) \mid (m(\tau_k) \subset \{m(\phi_i)\}) \wedge (m(\tau_k) \subset \{m(\varphi_j)\})\}, \quad (8)$$

where $i, j, k \in I^+$, $m(\phi_i)$ and $m(\varphi_j)$ are extensions of all the sub-concept granules of $(\phi, m(\phi))$ and $(\varphi, m(\varphi))$.

**Example 6**: Considering Figure 4(a) and Figure 4(b), Since the bottom concept granule of these two structures are all [Theory], we can use intersection operation to obtain a new granular knowledge structure, as Figure 4(c), which shows a partial structure that Figure 4(a) and Figure 4(b) both have. Since it appears in the analysis results of both proceedings, the partial structure may reflect hot research topics in the Rough Sets community.

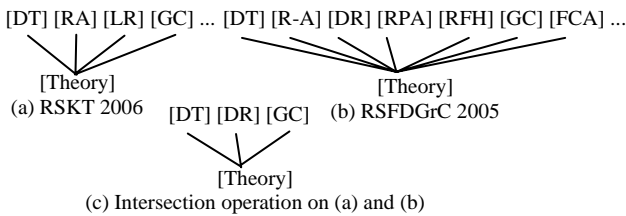

Fig. 4: Intersection operation on GKS.

**Definition 10**: (Difference Operation) Considering two concept granules $(\phi, m(\phi))$ and $(\varphi, m(\varphi))$ from two granular knowledge structures, if and only if $\phi = \varphi$, the difference operation, denoted as $(\phi, m(\phi)) - (\varphi, m(\varphi))$, can be used to find the difference among these two granular knowledge structures. The extensions of sub-concept granules in the new structure can be denoted as:

$$\{m(\tau_k) \mid (m(\tau_k) \subset \{m(\phi_i)\}) \wedge (m(\tau_k) \not\subset \{m(\varphi_j)\})\}, \quad (9)$$

where $i, j, k \in I^+$, $m(\phi_i)$ and $m(\varphi_j)$ are extensions of all the sub-concept granules of $(\phi, m(\phi))$ and $(\varphi, m(\varphi))$.

**Example 7**: Figure 5(a) and Figure 5(b) are granular knowledge structures representing related theory of Rough Sets based on proceedings of RSFDGrC 2005 and RSKT 2006. Through the difference operation on these two structures, we get a new structure, as shown in Figure 5(c), which shows related theories that Figure 5(a) has while Figure 5(b) doesn't have, namely, Logic and Reasoning, and Rough Approximation. This operation helps us to find the unique topics of a proceeding or a book, which others may don't contain.

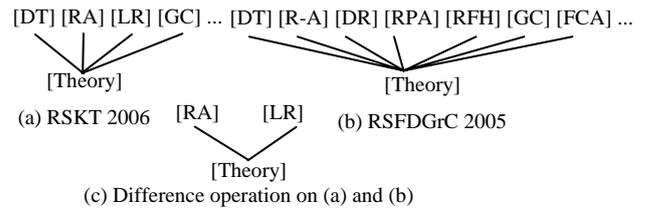

Fig. 5: Difference operation on GKS.

**Definition 11**: (Product Operation) Granular knowledge structures can be obtained based on partial ordered relations on a concept granule and its corresponding sub-concept granules. Let $(\phi, m(\phi))$ and $(\varphi, m(\varphi))$ be two concept granules. The extensions of their sub-concept granules, which are one level finer than them, can be denoted as:

$$m(\phi) = m(\phi_1) \cup m(\phi_2) \cup \ldots \cup m(\phi_n),$$
$$m(\varphi) = m(\varphi_1) \cup m(\varphi_2) \cup \ldots \cup m(\varphi_p),$$

where $n, p \in I^+$. Through the product operation, denoted as $(\phi, m(\phi)) \times (\varphi, m(\varphi))$, one can obtain a set of new concept granules.

$$\{m(\phi_i \wedge \varphi_j) \mid i = 1, \ldots, n; j = 1, \ldots, p\}, \quad (10)$$

where $m(\phi_i \wedge \varphi_j)$ is the extension of a co-defined concept granule by two super-concept granules. As shown in upper description, granular knowledge structures through product operation are produced by formula conjunction. If $(\phi, m(\phi))$ and $(\varphi, m(\varphi))$ share the same attribute and attribute value, by generalization operation, a super-concept granule of the two concept granules is added to show the connections between them for the produced granular knowledge structures.

**Example 8**: As shown in Figure 6, Let [Theory] and [Application Domain] be two concept granules, for simplicity, we choose sub-concept granules which are induced based on equality relation on $V_{Theory}$ = *{LR, RA, DR}* and $V_{ApplicationDomain}$=*{IR, MS, IS}* to produce a granular knowledge structure based on product operation. Since [Theory] and [Application Domain] share the attribute "Discipline" and the attribute value "Rough Sets", by generalization operation, the concept granule [Rough Sets] is added to show the connection between [Theory] and [Application Domain].

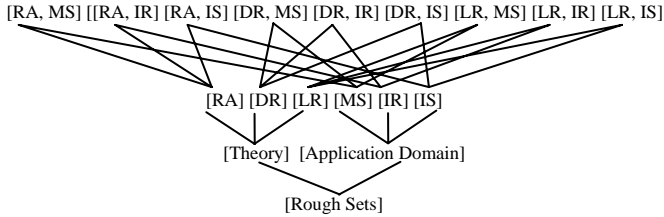

Fig. 6: Product operation on GKS.

The concrete meaning of this granular knowledge structure is as follows: in the bottom level, we just can conclude that these papers are about Rough Sets. In the second level, papers are categorized by "Theory" and "Application Domain". In the third level, they are classified by concrete values of "Theory" or "Application Domain". In the fourth level, the extension of each concept granule corresponds to a group of papers which are about an application domain and meanwhile use a related theory.

In granular knowledge structures induced by product operation, each level represents the concept granule in a certain degree of granularity. Different levels of concept granules form a partial ordering. The hierarchical structures describe the integrated whole of a web of concept granules from a very high level of abstraction to the very finest details.

As shown in upper examples, each granular knowledge structure provides a different understanding for a knowledge source. Granular knowledge structures may be much more complex than presented ones, but may be generated based on upper operations.

## 4 Using Granular Knowledge Structures

Reif and Heller argue that "effective problem solving in a realistic domain depends crucially on the content and structure of the knowledge about the particular domain" [2]. Hence, the use of granular knowledge structures could help one solve problems. Selections and switches on levels and views are two possible practical strategies on how to use granular knowledge structures.

### 4.1 Level selection and switch

In order to get detailed understanding of a granular knowledge structure, one may not only view it as an integrated whole, but also need to investigate concept granules among levels. For concrete tasks, some specific levels can be selected. Switching among those levels help users understand the knowledge source and meet their needs in different level of granularities [8]. In a granular knowledge structure, zoom-in operation and zoom-out operation are used for switching among levels.

**Definition 12**: (Zoom-in Operation) Let $\{(\phi_i, m(\phi_i)) \mid i \in I^+\}$ be a set of concept granules in the same level of a granular knowledge structure, and $\{(\phi_{11}, m(\phi_{11})), \ldots (\phi_{np}, m(\phi_{np})) \mid n, p \in I^+\}$ be the set of sub-concept granules one level finer than $\{(\phi_i, m(\phi_i)) \mid i \in I^+\}$. The zoom-in operation maps the set of concept granules from a coarser level to the one which are one level finer than it, following the notation in [17], it can be denoted as:

$$\omega((\phi_i, m(\phi_i))) = (\phi_{11}, m(\phi_{11})), \ldots (\phi_{np}, m(\phi_{np})) \quad (11)$$

**Definition 13**: (Zoom-out Operation) The inverse operation of zoom-in operation is zoom-out operation, denoted as $\omega^{-1}$, which maps the set of concept granules from a finer level to the one which are one level coarser than it. Let $i, n, p \in I^+$, it can be denoted as:

$$\omega^{-1}((\phi_{11}, m(\phi_{11})), \ldots (\phi_{np}, m(\phi_{np}))) = (\phi_i, m(\phi_i)) \quad (12)$$

Complex switches among levels are implemented by composition of zoom-in operation and zoom-out operation.

### 4.2 View selection and switch

It is emphasized that people with different background knowledge and purpose will have different understanding when learning from the same knowledge source [3]. For the same knowledge source, different views may induce different granular knowledge structures, and one can get different understandings of the knowledge source through each of them. In upper sections of this paper, we examined concrete examples in the field of scientific literature, and we provide different granular knowledge structures based on various operations. Each granular knowledge structure shows a unique understanding of the papers in those two proceedings. Even for the same granular knowledge structure, one can get different understanding when different viewpoint is selected [3].

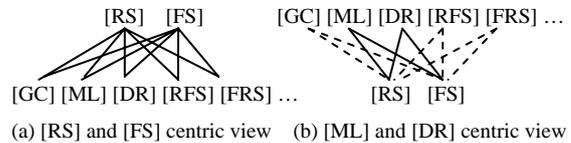

(a) [RS] and [FS] centric view   (b) [ML] and [DR] centric view

Fig. 7: View switch according to different background Knowledge.
RS: Rough Sets, FS: Fuzzy Sets, ML: Machine Learning, RFS: Rough Fuzzy Sets, FRS: Fuzzy Rough Sets.

**Example 9**: Figure 7(a) shows an analysis of the 1st-4th China National Rough Sets and Soft Computing Conference proceedings from the viewpoint of main related fields, namely, Rough Sets, Fuzzy Sets. The concept granules [RS] and [FS] form a partial ordering with their sub-concept granules respectively. We can conclude that "data reduction" and "machine learning" are two related fields for both Rough Sets and Fuzzy Sets. This piece of knowledge indicates that researchers on Rough Sets and Fuzzy Sets can work on "data reduction" and "machine learning". If we switch to another view to investigate the picture (as in Figure 7(b)), [ML] and [DR] are all related to [RS] and [FS], which indicates that both Rough Sets and Fuzzy Sets are approaches to "data reduction" and "machine learning", which tells us that for data reduction and machine learning researchers, "Rough Sets" and "Fuzzy Sets" may be two possible theoretical methods for their research.

## 5 Conclusion

In this paper, we provide our understanding on interpreting knowledge from the viewpoint of granular computing and examine different granular knowledge structures based on various operations. Different granular knowledge structures provide different views of the knowledge source. Each view provides a unique understanding.

Granular knowledge structures provide understandings of knowledge in two aspects. Firstly, through representation of a granular knowledge structures based on concept granules and their relations, they provide an understanding of knowledge from the set theoretic and logic point of view. Secondly, through visualized structures, they provide an easily acceptable way for users to understand knowledge. In fact, the visualized structure shows how those set theoretic and logical representations are organized [12].

Examples in this paper has shown some impact of granular knowledge structures in helping users understand the knowledge source from multiple levels and multiple views. Considering its characteristics and expressiveness, granular knowledge structures may have wider use in other fields related to human and machine intelligence.

## Acknowledgements

This work is supported by National Natural Science Foundation of China research program (No. 60673015), the Open Foundation of Beijing Municipal Key Laboratory of Multimedia and Intelligent Software Technology. The authors would like to thank Professor Yiyu Yao and Lina Zhao for their constructive discussion on this paper.

## References

[1] A.M. Collins, M.R. Quillian. "Retrieval time from semantic memory", *Journal of Verbal Learning and Verbal Behavior*, **volume 8**, pp. 240-248, (1969).

[2] F. Reif, J. Heller. "Knowledge structure and problem solving in physics", *Educational Psychologist*, **volume 17**, pp. 102-127, (1982).

[3] J.D. Bransford, A.L. Brown, and R.R. Cocking (eds). How People Learn: Brain, Mind, Experience, and School, National Academy Press, (2000).

[4] J.F. Sowa. Conceptual Structures, Information Processing in Mind and Machine, Addison-Wesley Publishing Company, Inc., Massachusetts, (1984).

[5] J. Larkin, H. Simon, "Why a diagram is (sometimes) worth ten thousand words", *Cognitive Science*, **volume 11**, pp. 65-99, (1987).

[6] L.A. Zadeh. "Towards a theory of fuzzy information granulation and its centrality in human reasoning and fuzzy logic", *Fuzzy Sets and Systems*, **volume 19**, pp. 111-127, (1997).

[7] M. Buchheit, F.M. Donini, and A. Schaerf. "Decidable reasoning in terminological knowledge representation systems", *Journal of Artificial Intelligence Research*, **volume 1**, pp.109-138, (1993).

[8] M. Minsky. The Emotion Machine: Commonsense Thinking, Artificial Intelligence, and the Future of the Human Mind, Simon & Schuster, Inc., (2006).

[9] P. James. "Knowledge graphs", In: R.P. van de Riet, R.A. Meersman (eds). Linguistic Instruments in Knowledge Engineering, Elsevier Science Inc., (1992).

[10] W. Goffman. "Mathematical approach to the spread of scientific ideas-the history of mast cell research", *Nature*, **volume 212**, pp. 449-452, (1966).

[11] W. Goffman. "A mathematical method for analyzing the growth of a scientific discipline". *Journal of the Association for Computing Machinery*, **volume 18(2)**, pp. 173-185, (1971).

[12] W. Homenda, "Granular structures: the perspective of knowledge representation", Proceedings of the joint 9th IFSA World Congress and 20th NAFIPS International Conference, **volume 3**, pp. 1682-1687, (2001).

[13] Y.H. Chen, Y.Y. Yao. "A multiview approach for intelligent data analysis based on data operators", *Information Sciences*, **volume 178(1)**, pp. 1-20, (2008).

[14] Z. Pawlak, Rough Sets, Theoretical Aspects of Reasoning about Data. Kluwer Academic Publishers, Dordrecht, (1991).

[15] Y.Y. Yao, N. Zhong, "Granular computing using information tables", In: T.Y. Lin, Y.Y. Yao, and L.A. Zadeh (eds). Data Mining, Rough Sets and Granular Computing, Physica-Verlag, pp. 102-124, (2002).

[16] Y.Y. Yao. "Concept formation and learning: a cognitive informatics perspective", Proceedings of the 3rd IEEE International Conference on Cognitive Informatics, IEEE Press, pp. 42-51, (2004).

[17] Y.Y. Yao, "A partition model of granular computing". *LNCS Transactions on Rough Sets I*, LNCS 3100, Springer, pp. 232-253, (2004).

[18] Y.Y. Yao, Y. Zeng, and N. Zhong. "Supporting literature exploration with granular knowledge structures", Proceedings of the 11th International Conference on Rough Sets, Fuzzy Sets, Data Mining and Granular Computing, LNAI 4482, Springer, pp. 182-189, (2007).

[19] Y.Y. Yao. "The art of granular computing". Proceedings of the International Conference on Rough Sets and Emerging Intelligent Systems Paradigms, LNAI 4585, Springer, pp. 101-112, (2007).

[20] Y.Y. Yao, B. Zhou. "A logic language of granular computing", Proceedings of the 6th IEEE International Conference on Cognitive Informatics, IEEE Press, pp. 178-185, (2007).